\documentclass{article}
\PassOptionsToPackage{numbers,sort&compress,square}{natbib}
\usepackage[preprint]{neuripsformat} 

\usepackage[utf8]{inputenc}                             
\usepackage[T1]{fontenc}         
\usepackage[breaklinks,colorlinks]{hyperref} 
\usepackage{url}                                        
\usepackage{booktabs}                                   
\usepackage{amsfonts}
\usepackage{amsmath}
\usepackage{mathtools}
\usepackage{nicefrac}                                    
\usepackage{microtype}                                  
\usepackage{xcolor}         
\usepackage{graphicx}
\usepackage{tabularray}
\usepackage{bm}
\usepackage{icomma}
\usepackage{multirow}
\usepackage[labelformat=empty]{subfig}
\usepackage{cleveref}
\usepackage{printlen}
\usepackage{svg}
\usepackage{tikz}
\usepackage{enumitem}
\usetikzlibrary{matrix}

\newcommand{\x}{\mathbb{\mathbf{x}}}

\newcommand{\z}{\mathbb{\mathbf{z}}}
\newcommand{\norm}[1]{\left\Vert #1 \right\Vert}

\title{
Understanding and Visualizing Droplet Distributions in Simulations of Shallow Clouds
}

\author{
Justus C. Will$^1$,
Andrea M. Jenney$^{1,2}$,
Kara D. Lamb$^{3}$,
Michael S. Pritchard$^{1, 4}$, \\\bfseries
Colleen Kaul$^5$, 
Po-Lun Ma$^5$,
Kyle Pressel$^5$,
Jacob Shpund$^5$, \\\bfseries
Marcus van Lier-Walqui$^3$,
Stephan Mandt$^1$
\\
$^1$UCI, $^2$OSU, $^3$Columbia, $^4$NVIDIA $^5$PNNL,
}

\begin{document}
\maketitle

\begin{abstract}

Thorough analysis of local droplet-level interactions is crucial to better understand the microphysical processes in clouds and their effect on the global climate.
High-accuracy simulations of relevant droplet size distributions from Large Eddy Simulations (LES) of bin microphysics challenge current analysis techniques due to their high dimensionality involving three spatial dimensions, time, and a continuous range of droplet sizes.
Utilizing the compact latent representations from Variational Autoencoders (VAEs), we produce novel and intuitive visualizations for the organization of droplet sizes and their evolution over time beyond what is possible with clustering techniques.
This greatly improves interpretation and allows us to examine aerosol-cloud interactions by contrasting simulations with different aerosol concentrations.
We find that the evolution of the droplet spectrum is similar across aerosol levels but occurs at different paces. This similarity suggests that precipitation initiation processes are alike despite variations in onset times.

\end{abstract}

\section{Introduction}
\label{sec:introduction}

Understanding and accurately representing cloud processes in numerical
models is crucial for improving weather and climate predictions. Cloud droplets and their size distributions play a significant role in various atmospheric phenomena, such as radiation and precipitation initiation, making their characterization essential. However, complete simulation of these processes remains prohibitive in numerical models of the atmosphere due to their high complexity and small physical scale. Instead, cloud physics are represented through parameterizations, greatly simplified processes that often rely on assumptions about the shape of cloud droplet distributions over volumes and the size of a numerical model’s grid cell, which remain largely under-verified using observations.

Numerical simulations with more sophisticated cloud microphysics parameterizations (i.e., relying on fewer or no assumptions about the shape of droplet distributions) are used to inform the next generation of cloud physics models.
The motivation for this study arises from the need to efficiently and effectively summarize the simulated droplet distributions from a pioneering set of Large Eddy Simulations of shallow clouds. These simulations provide droplet bin masses for every grid cell at a relatively high temporal frequency. While previous studies have employed clustering algorithms on observed droplets for this task (e.g., Allwayin et al. \cite{allwayin2022automated}), these methods pose challenges for our data due to their sheer size. We are inspired by recent advancements in machine learning, particularly Variational Autoencoders (VAEs), which have shown promise in capturing patterns in complex climate datasets while preserving physical interpretability \cite{mooers2021analyzing, mooers2023comparing, behrens2022nonlinear}.
Our main contributions incldue:
%
\begin{itemize}[noitemsep,topsep=0pt,parsep=0pt,partopsep=0pt]
    \item We propose a new way to visualize high-dimensional, spatio-temporal droplet size distributions by a VAE-based approach, representing droplet distributions through color spectra.
     \item We characterize the transition of droplet distributions from ambient to precipitating.
    \item Our analysis confirms that aerosol concentrations may delay precipitation onset.
\end{itemize}

\subsection{LES Simulations}
\label{sec:data}

\begin{figure}
    \centering
    \subfloat[\centering $t = 2$h]{{\includegraphics[width=.33\textwidth]{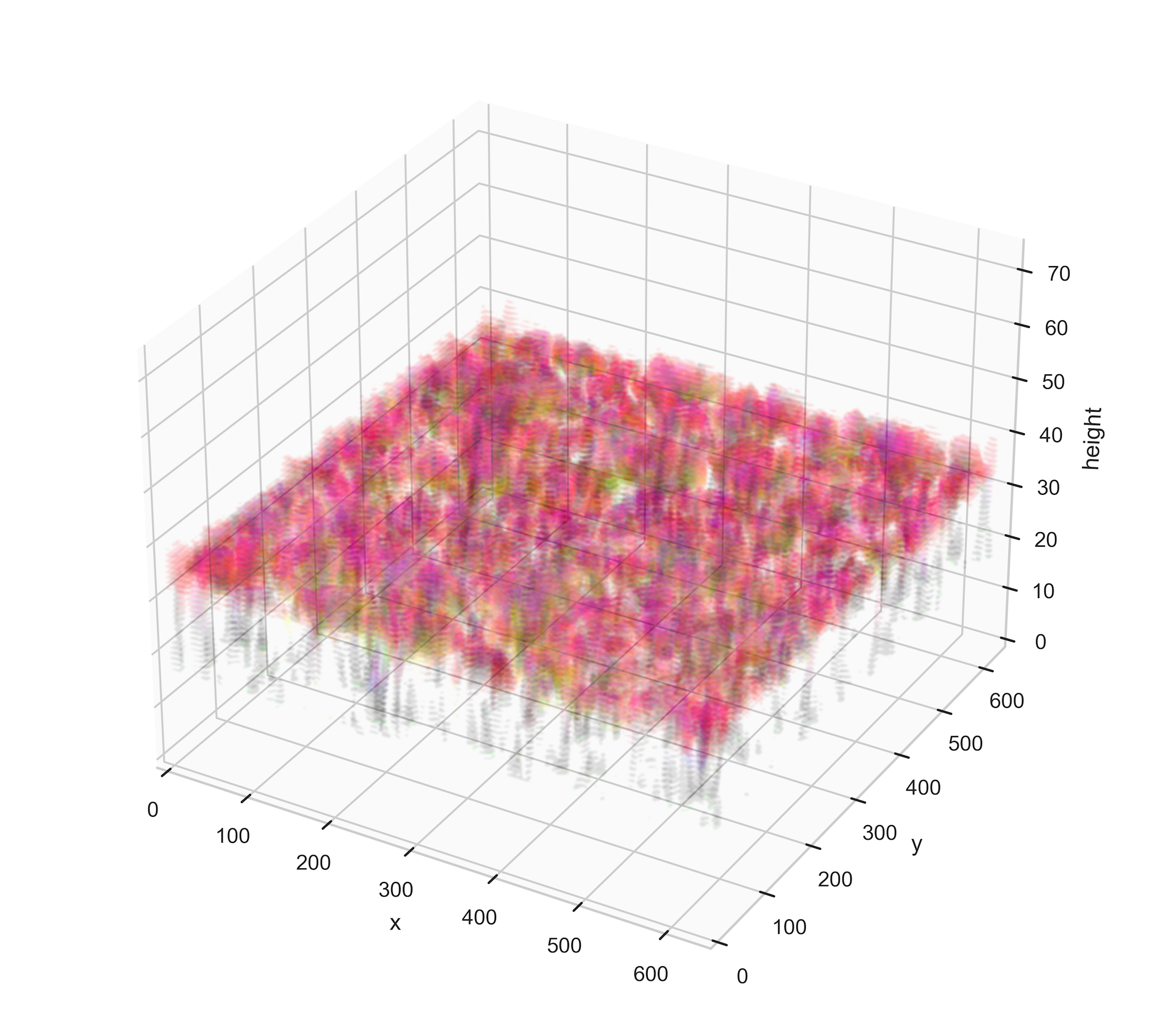}}}
    \subfloat[\centering $t = 4$h]{{\includegraphics[width=.33\textwidth]{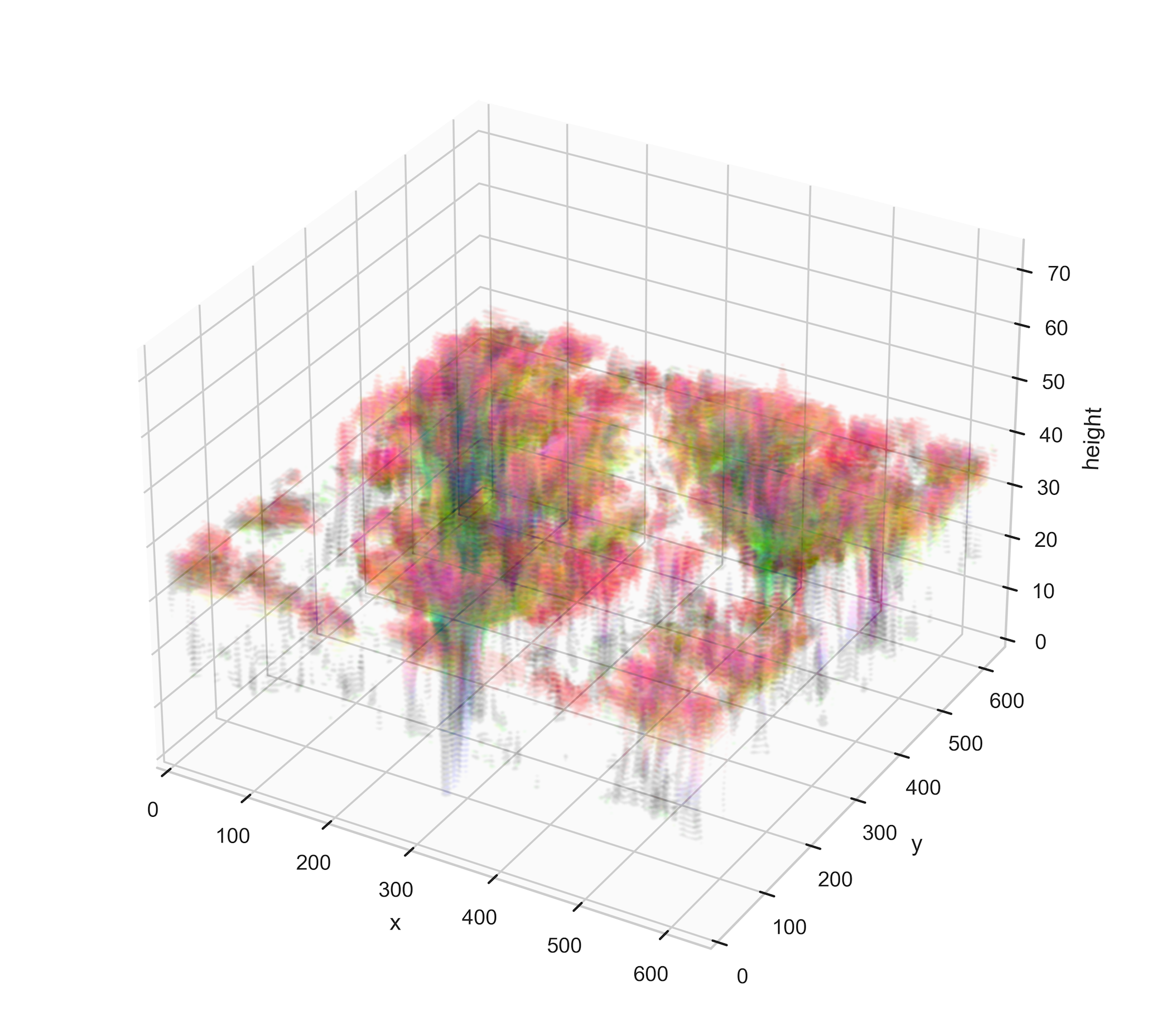}}}
    \subfloat[\centering $t = 6$h]{{\includegraphics[width=.33\textwidth]{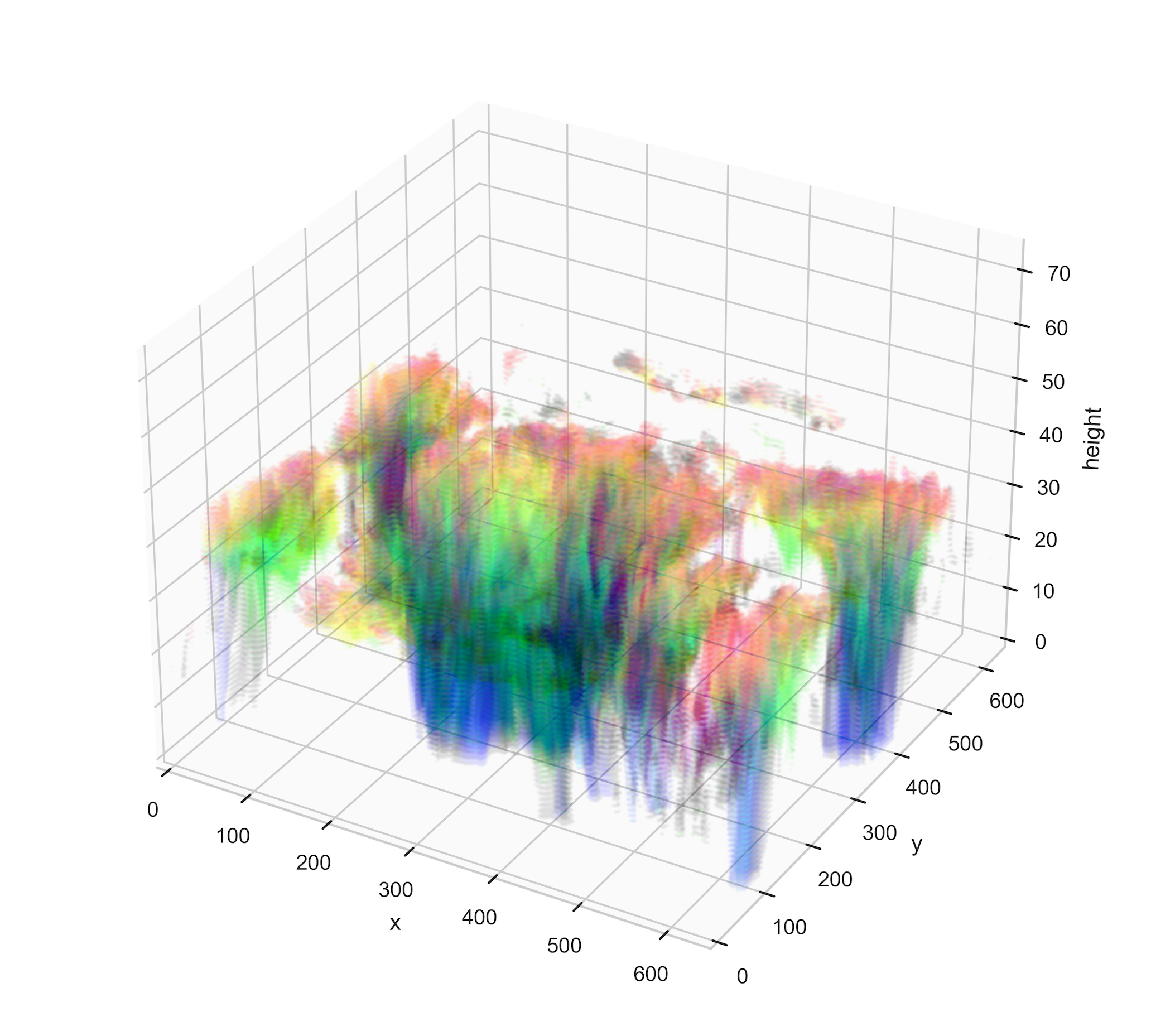}}}
    \caption{The time evolution of the spatial organization of droplet size distributions in simulations of shallow clouds at a base aerosol level. Color represents latent space location as defined in \Cref{sec:visualization} and thus indicates distribution characteristics. Precipitating regions appear only at later times.}
    \label{fig:clouds}
\end{figure}

Several LES simulations were run under different meteorological conditions using the PINACLES model \cite{pressel2021developing}.
For the sake of brevity, we focus our analysis on simulations of warm clouds in the \emph{trade cumuli} regime based on the \emph{ATEX} campaign \cite{ATEXsim}. We note that our methodology is applicable to a broader set of simulations which also includes, for example, \emph{nocturnal stratocumulus} based on the \emph{DYCOMS} campaign \cite{DYCOMSsim}.
In all simulations, microphysical processes are resolved using the \emph{Fast Spectral Bin Microphysics version 2} \cite{Shpund2019}, which defines cloud droplet size distributions (DSDs) using $33$ mass-doubling bins up to a maximum diameter of $6.5$ mm. They are run for 8 hours of simulated time with an internal timestep of roughly 1 second. Three-dimensional snapshots of the $25.6\!\times\!25.6\!\times\!3$ km doubly periodic domain (with a grid resolution of $40$m) are taken every $10$ minutes. Three separate simulations are run at half, base, and double the published aerosol concentrations as prescribed for the RICO study \cite{RICOsim}, allowing us to isolate and analyze cloud-aerosol interactions.
For computational efficiency, we discard all DSDs associated with clear air, i.e. whose summed mixing ratio (mass of liquid per unit of dry air) falls below a threshold of $10^{-5}$. Furthermore, whenever DSDs are used as input to neural networks, their summed mixing ratio is normalized to $1$. This allows for faster and more stable learning and avoids giving less importance to DSDs with less mass. 

\section{Variational Autoencoder and Learned Representation}
\label{sec:vae}

A recent study by Lamb et al. \cite{lamb2023reduced} suggests that droplet collision-coalescence, which is the most important processes governing the time evolution of DSDs, has an inherent dimensionality of $3$. This motivates the use of a learned $3$-dimensional representation, which empirically captures all important characteristics, even in our more complex setting including spatial interaction and aerosols.
Specifically, we use Variational Autoencoders (VAEs) \cite{kingma2014auto}, which are generative latent variable models that can be fit to data $\mathcal{D} = \{\x_1, \dots, \x_n\}$, learning both a low-dimensional representation of data samples and enabling controlled generation of new data. In contrast to non-stochastic autoencoders this allows us to find more robust representations that better generalize to new samples and to quantify data variability and model uncertainty.
To this end, we define a joint likelihood over data $\x$ and a lower-dimensional latent variable $\z$, where $\z$ informs a complex conditional distribution $p_\theta(\x|\z)$ over the data domain -- in our case, a Gaussian distribution whose mean is parameterized by a feed-forward neural network (MLP) $\mu_\theta(\z)$ (the variational decoder).
To fit this model to data, we use amortized variational inference to minimize the negative evidence lower bound (NELBO) $\mathcal{L}_\theta(q)$, which uses a Gaussian approximation $q_\psi$ (with mean $g_\psi(\x_i)$ and $h_\psi(\x_i)$ parameterized using MLPs) to the posterior $p_\theta(\z|\x)$ to tightly bound the intractable negative marginal likelihood from above. This is equivalent to minimization with the loss
\begin{align*}
    \label{eq:nelbo}
  \mathcal{L}_{\theta, \psi}(\x) =
  \mathbb{E}_{\z\sim q_\psi} \left[ \frac{1}{2}\norm{\x - \mu_\theta(\z)}_2^2\right] + \beta \, \mathrm{KL}(q_\psi(\z) \,\|\, p(\z))
\end{align*}
so that suitable parameters $(\theta, \psi)$ can be found with stochastic gradient-based optimization techniques. 

\Cref{fig:latent} illustrates the latent space of our final model showing the point cloud of encoded latent representations for all DSDs across all time steps and aerosol levels. Encoded points close in this latent space directly correspond to DSDs with similar characteristics so that the spatial organization in the latent space meaningfully represents the inherent structure present in the data.
Specifically, we observe that the regions with high point density form a highly connected continuum, indicating the presence of a very continuous transition between DSDs of different characteristics, even in distribution space.
We identify a large, homogeneous, and roughly spherical region centered at zero that smoothly transitions into a separate narrow filament structure that traces a path with a sharp bend.

\section{Visualization and Insights}
\label{sec:visualization}

\begin{figure}
    \centering
    \centering
    \subfloat[a)\label{fig:latent}]{{\includegraphics[width=0.5\textwidth]{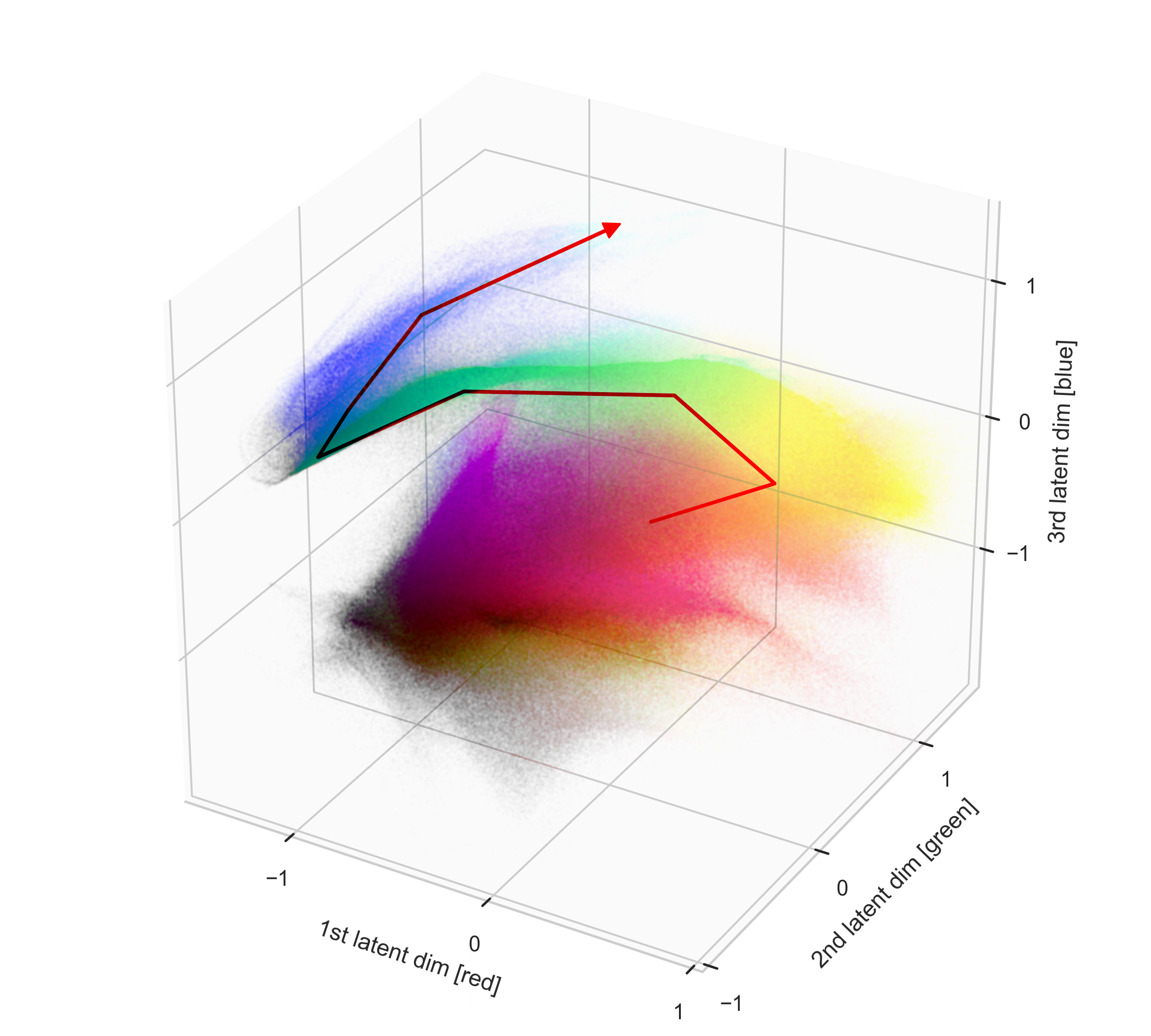}}}
    \subfloat[b)\label{fig:dsd}]{{\includegraphics[width=0.5\textwidth]{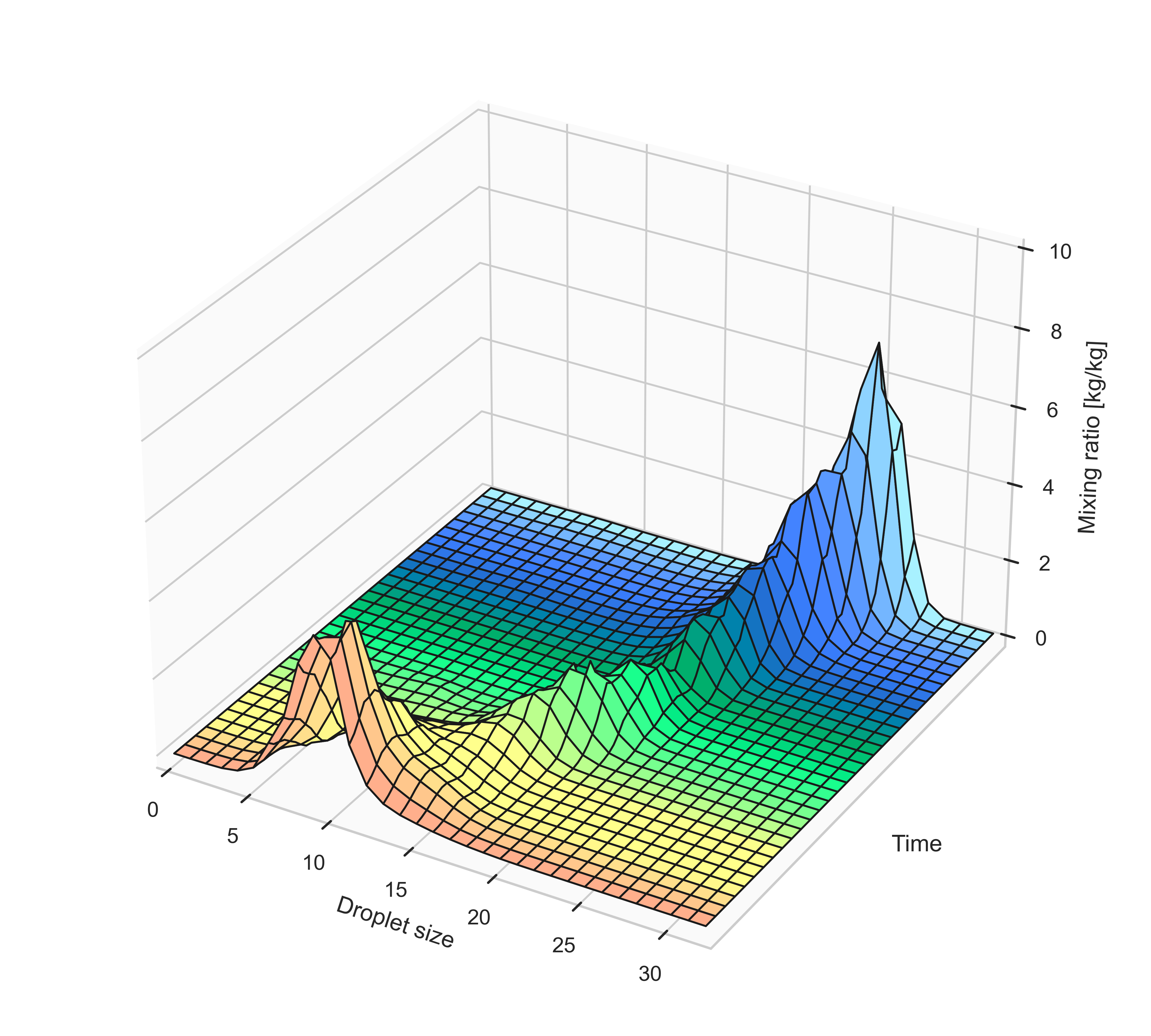}}}
    
    \caption{a) The joint VAE latent space over all time steps and aerosol levels. Color represents latent space location as defined in \Cref{sec:visualization}. The red arrow marks the pathway of precipitation, retrieved based on the latent space evolution through time. 
    b)
    Evolution of the droplet size distributions along the pathway, from ambient to precipitating distributions.}
    
\end{figure}

Representing the 3D latent space location with colors, we can assign continuous labels to different regions to permit a 1D \textit{interpretation} of latent space ``neighborhoods'' without any clustering or information loss. 
Specifically, we make use of the fact that color itself can be described using a three-dimensional spectrum and map the latent variable $z$ onto a color where the value in the first, second, and third latent dimension linearly corresponds to the amount of red, green, and blue in the color.
\Cref{fig:latent} shows each data point colored using this RGB representation.

\Cref{fig:clouds} shows the time evolution of DSDs in a simulation of clouds from the \emph{ATEX} meteorological case at the base aerosol level. Looking at the spatial organization allows us to better understand the role DSDs of different characteristics play.
We note that even as early as $2$ hours, well in advance of the occurrence of large cohesive shafts of large droplets extending down to the surface (i.e., precipitation), small pockets of yellow-to-green DSD form, which later become associated more with the precipitating regions, where rain seems to form a yellow-green-blue transition as the droplets get bigger and start to fall to lower altitudes. 

The emergence of precipitation regions is also clearly visible in the latent space, where the associated filament structure only appears at later time steps, when mass starts moving along the path as indicated in \Cref{fig:latent}. By tracing the retrieved path in the latent space and relating it back to associated distributions, we can get valuable insight about distribution transtitions along the path of precipitation.
Specifically, for each point on the latent space path, we average the $1000$ observed DSDs whose encoded representations are closest, in a Euclidean sense, to the point of interest. The obtained distribution evolution is shown in \Cref{fig:dsd} and characterized by a steady increase in droplet size, again confirming the close association with rainfall.
\begin{figure}
    \centering
    \begin{tikzpicture}
    \matrix (fig) [matrix of nodes, inner sep=0pt]{
        \includegraphics[width=.3\textwidth]{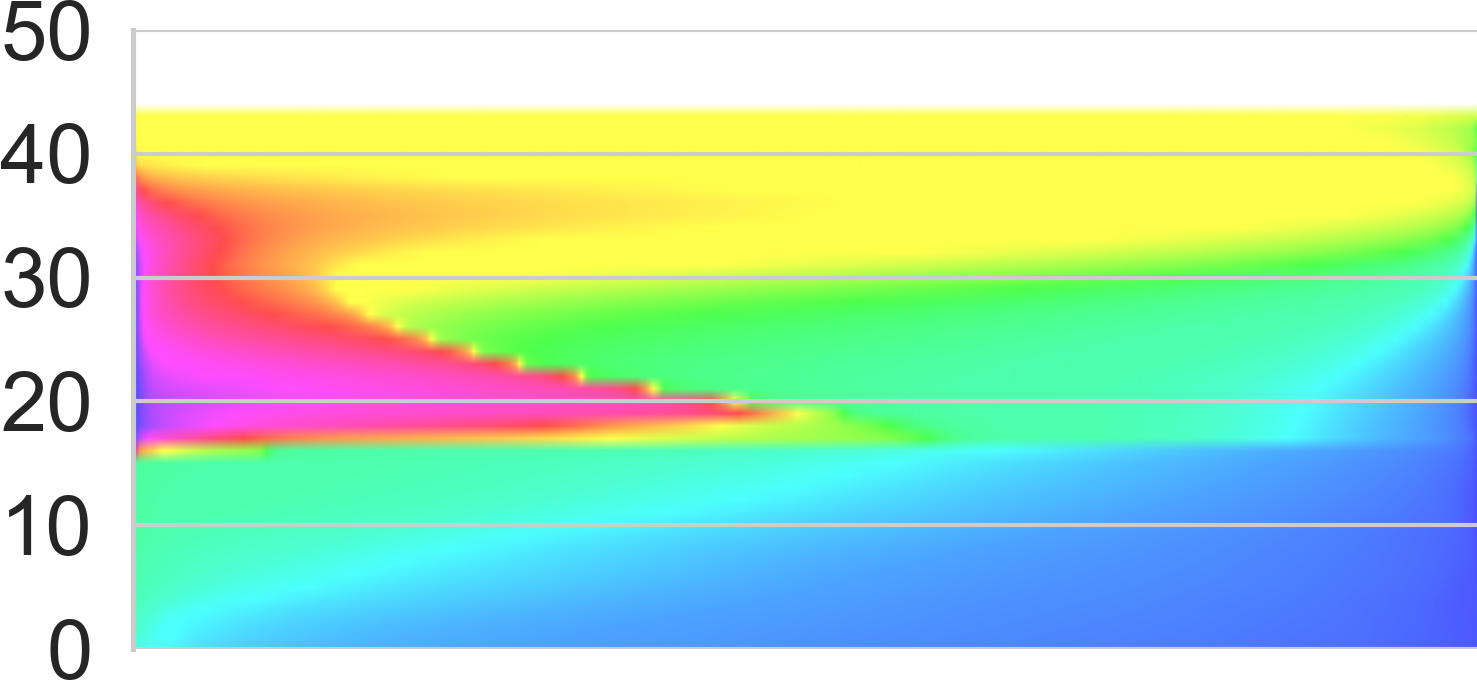}
        &
        \includegraphics[width=.3\textwidth]{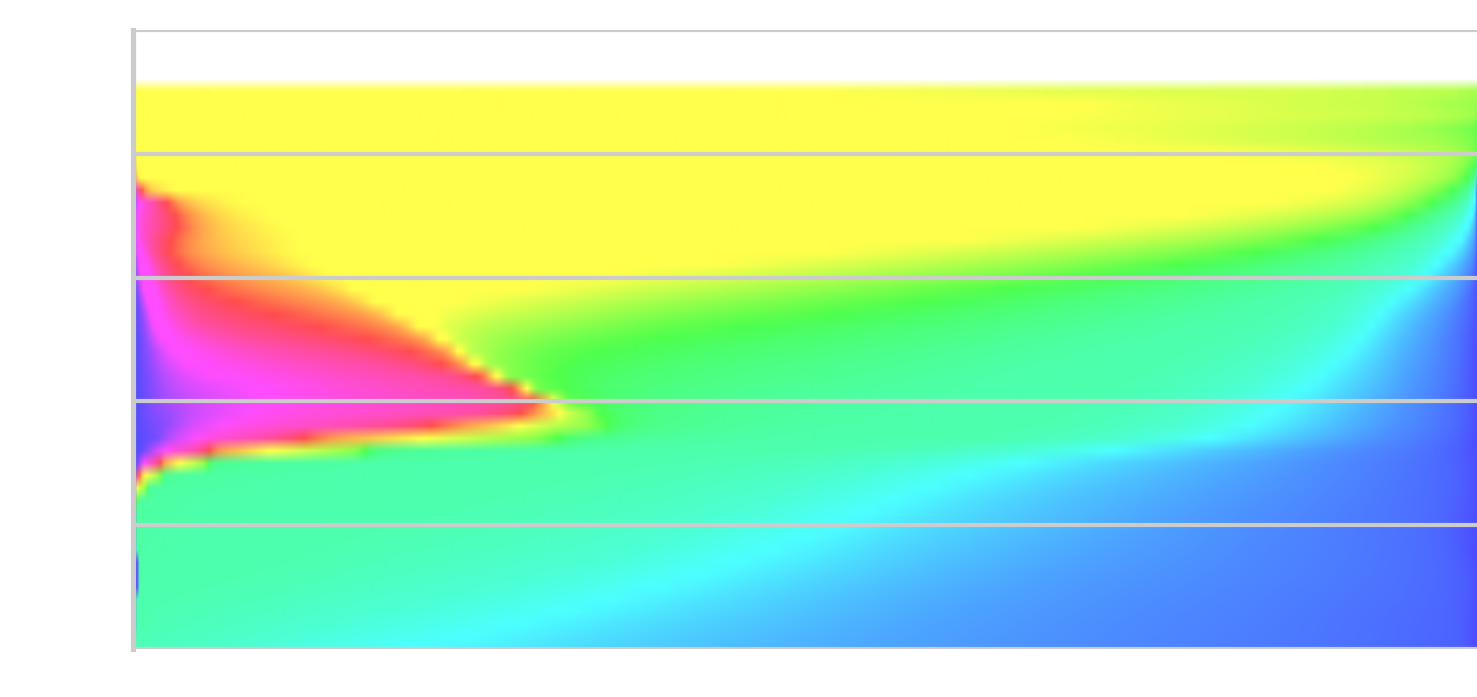}
        &
        \includegraphics[width=.3\textwidth]{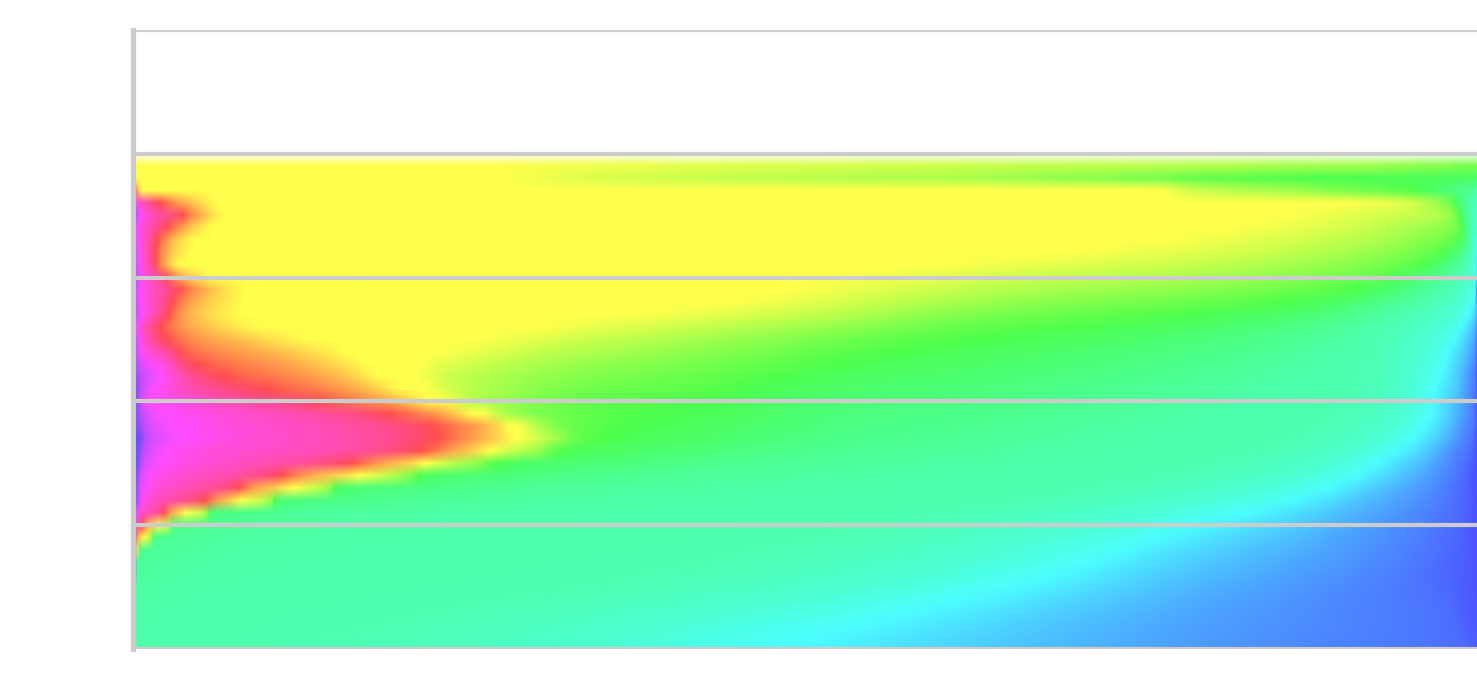}
        \\
        \includegraphics[width=.3\textwidth]{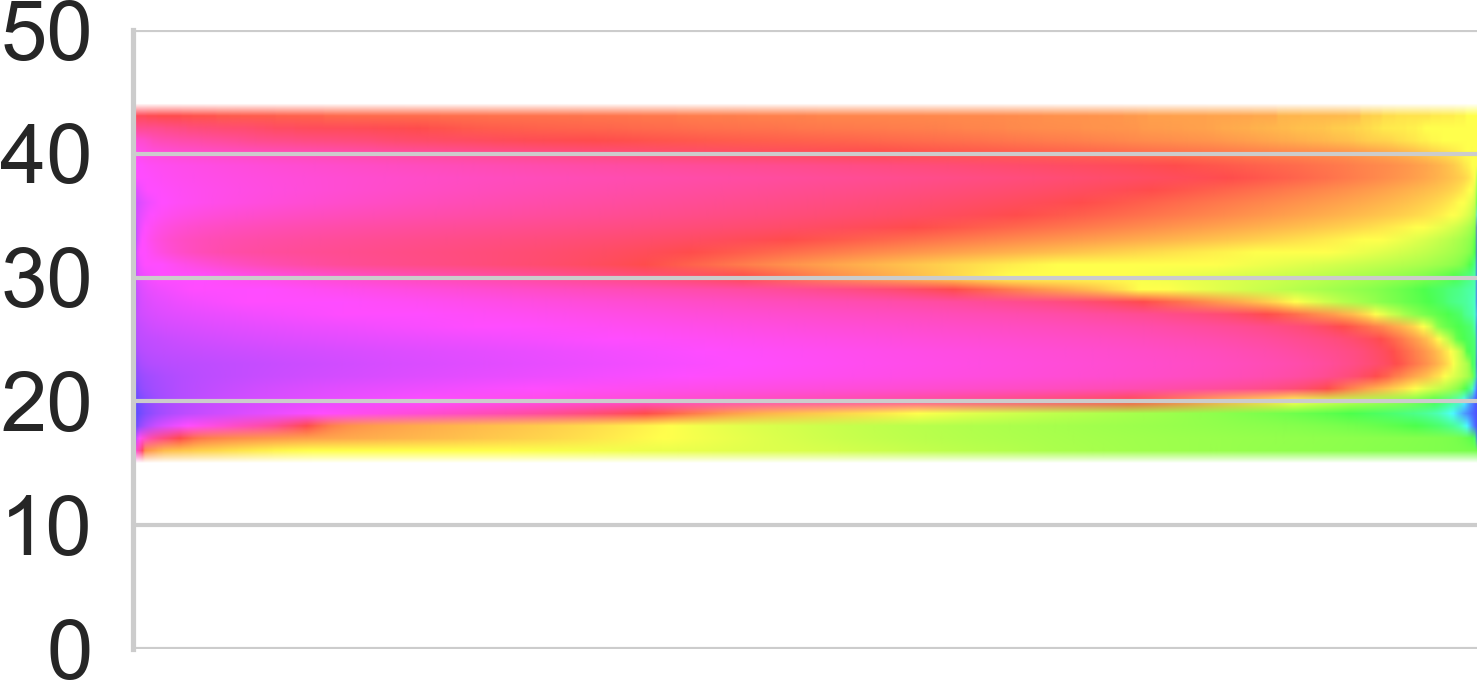}
        &
        \includegraphics[width=.3\textwidth]{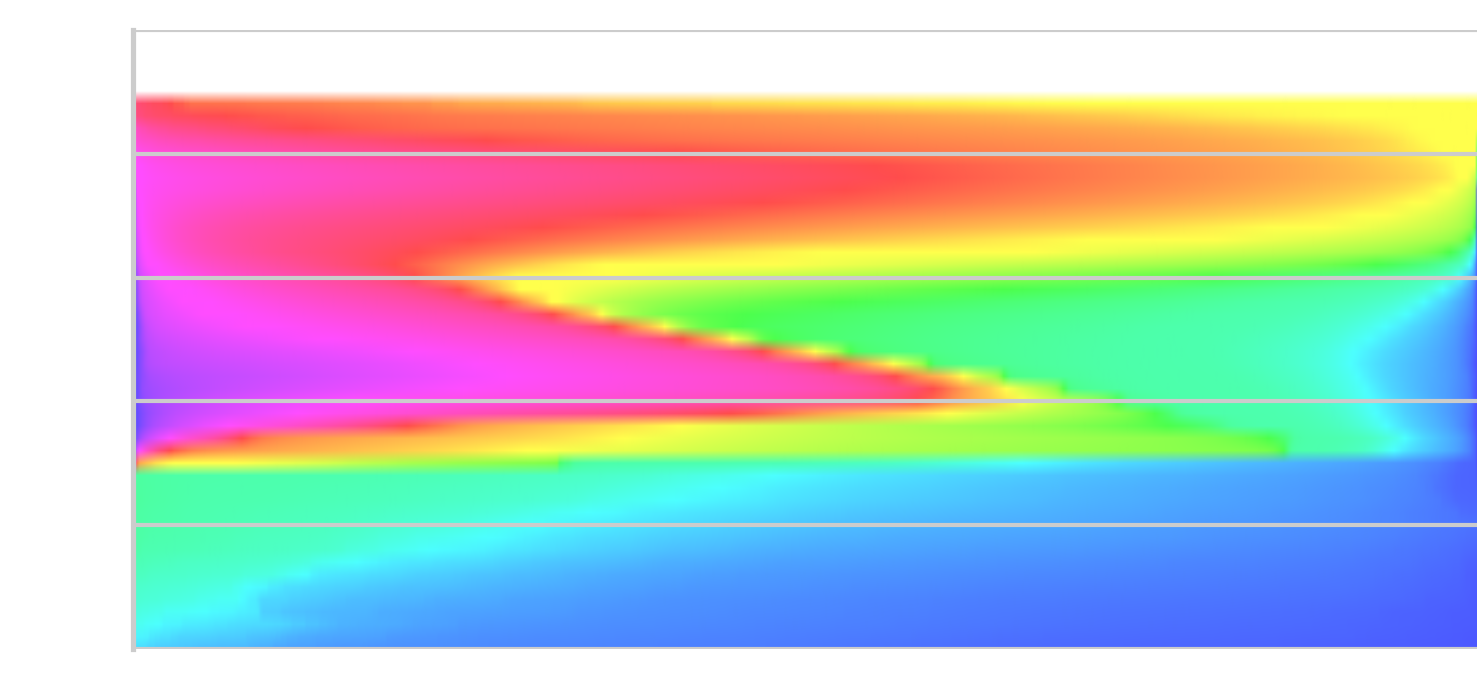}
        &
        \includegraphics[width=.3\textwidth]{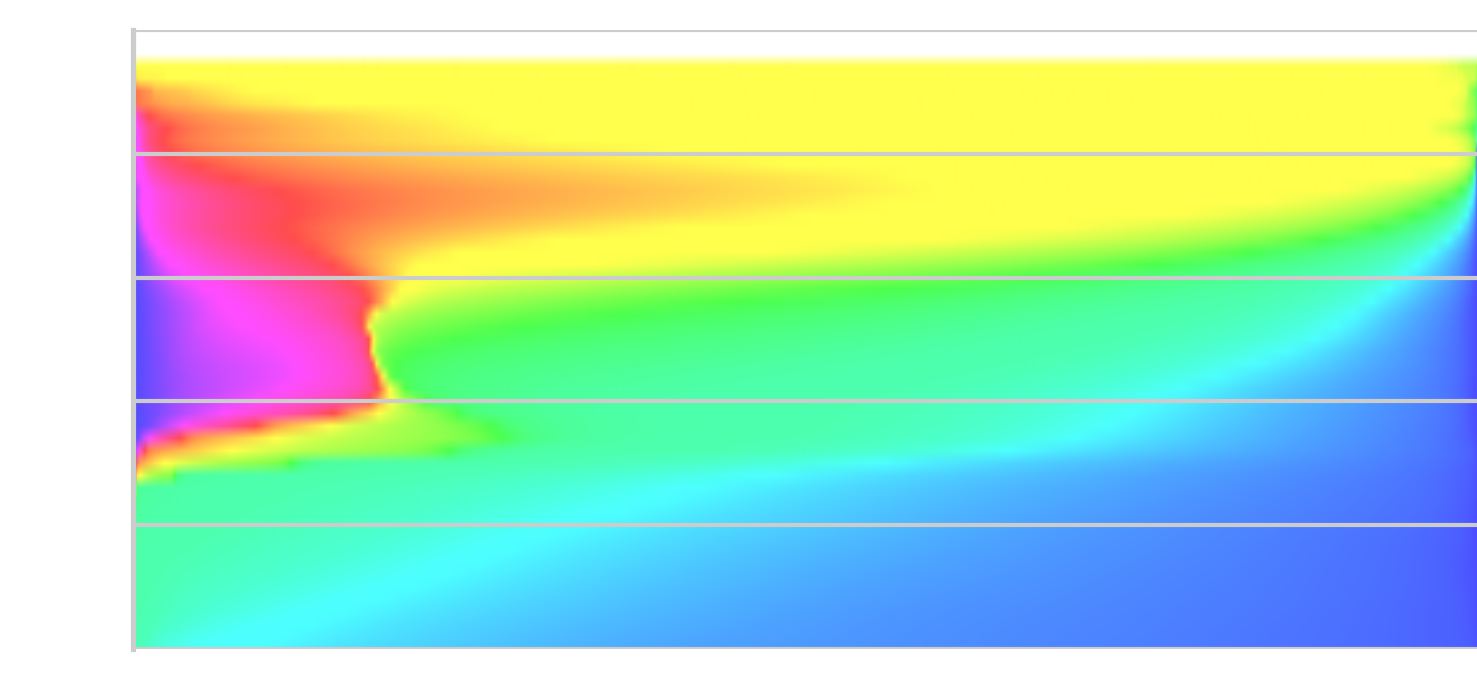}
        \\
        \includegraphics[width=.3\textwidth]{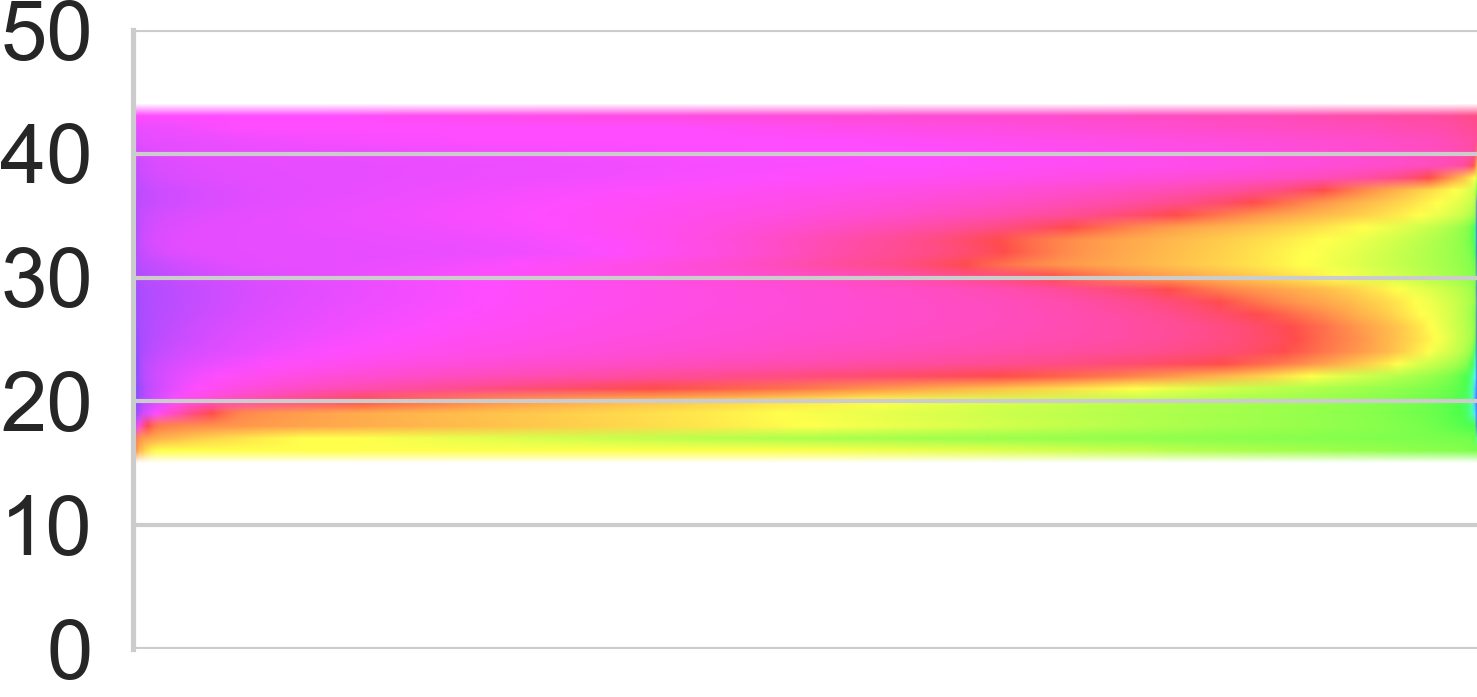}
        &
        \includegraphics[width=.3\textwidth]{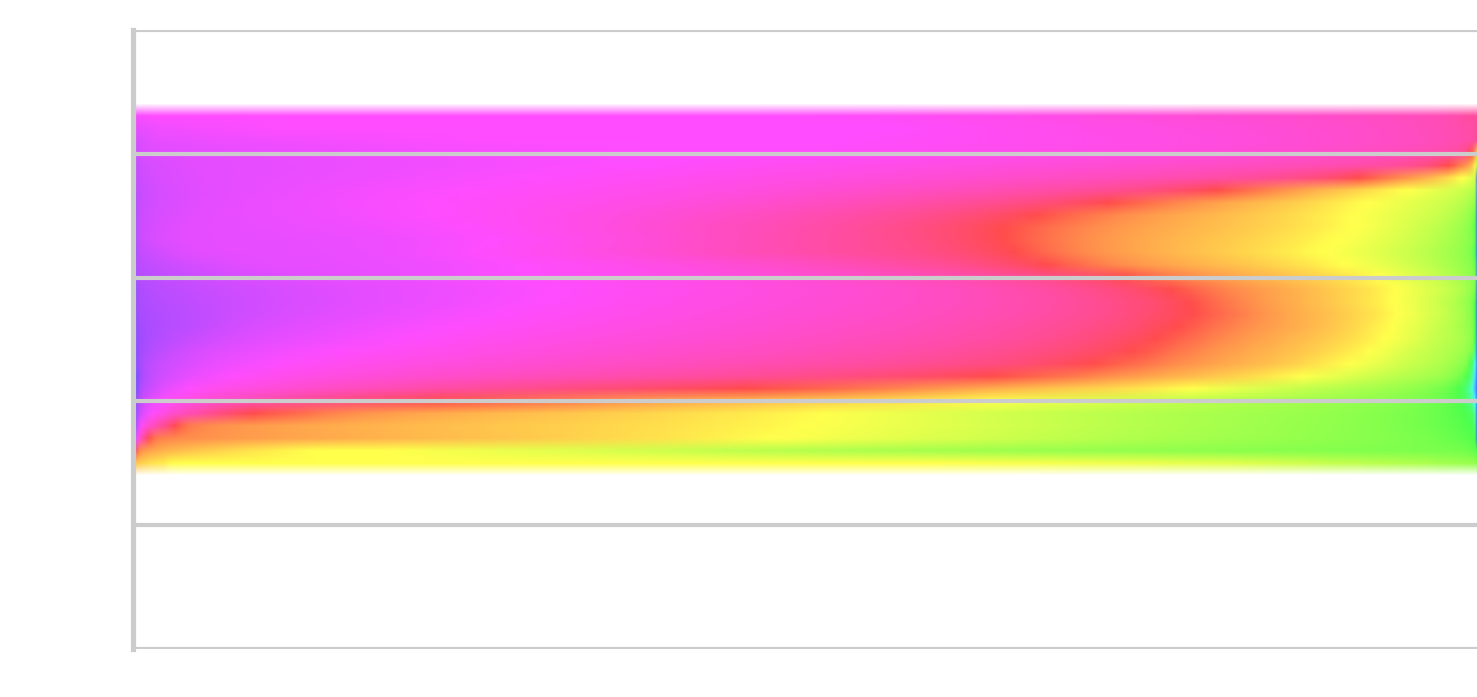}
        &
        \includegraphics[width=.3\textwidth]{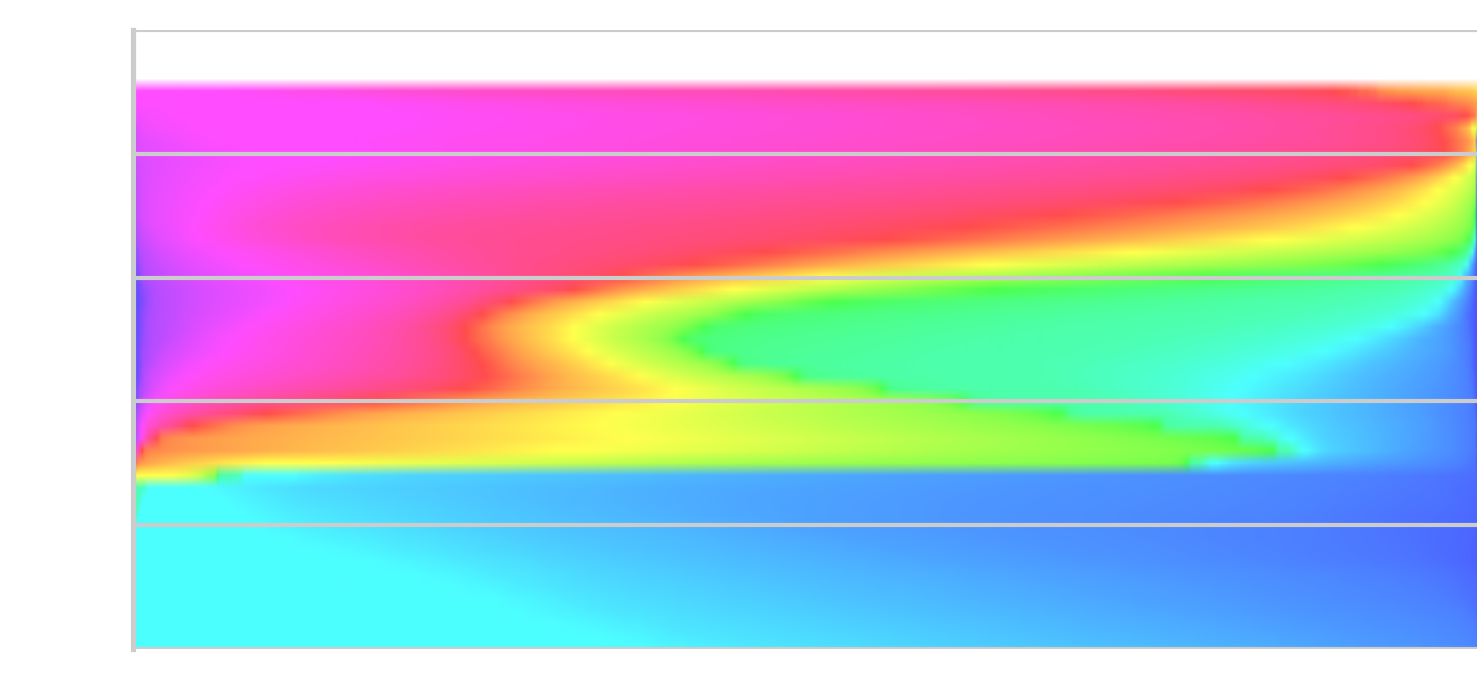}
        \\
    };
    \node at (fig-1-1.north) [above] {\hspace{1.4em} $t = 2$h};
    \node at (fig-1-2.north) [above] {\hspace{1.4em} $t = 4$h};
    \node at (fig-1-3.north) [above] {\hspace{1.4em} $t = 7$h};
    \node at (fig-2-1.west) [above, rotate=90] {altitude};
    \node at (fig-1-3.east) [below, rotate=90] {$0.5\times$};
    \node (m) at (fig-2-3.east) [below, rotate=90] {$1\times$};
    \node at (fig-3-3.east) [below, rotate=90] {$2\times$};
    \node at (m) [below, rotate=90] {base aerosols};
    \end{tikzpicture}
    \caption{The relative occurrence of specific droplet size distributions per height level. Color is normalized and represents latent space location as discussed in \Cref{sec:visualization}. Increasing levels of aerosols delay the precipitation onset.}
    \label{fig:proportions1d}
\end{figure}

Finally, we can summarize the state of the simulation at each time step by looking at what proportion of DSDs follow specific characteristics/colors. This allows us to pinpoint precipitation initiation to the moment when the presence of DSDs in precipitating regions, associated with green and blue colors, significantly increases at later simulation times. This analysis is similar to tracking distribution across clusters over time, only for the case of continuous labels using a full spectrum of colors instead.
\Cref{fig:proportions1d} shows, for each altitude level of the model, the relative occurrence of specific droplet size distributions at $2$, $4$, and $7$ hours for each aerosol concentration. Information about the horizontal position and vertical \textit{structure} is discarded. Specifically, we sort the set of all DSDs in the horizontal plane (for fixed aerosol, time step, and height) by hue. After normalization of saturation and brightness, this leaves us with a smooth color transition from violet/pink to blue colors that show proportionality while roughly following the transition from ambient DSDs with mainly small droplet sizes to DSDs associated with precipitation.

The composition plots in \Cref{fig:proportions1d} enable the fast summary of the state of the simulation and allow for insightful comparisons across different simulation conditions.
For aerosol concentration in particular, we note that an increase in aerosol concentration causes a delay in the onset of precipitation. At a base aerosol level, green and blue DSDs appear in significant amounts only after roughly $4$ hours. With less aerosols, this happens $2$ hours earlier and with more aerosols $3$ hours later. Delayed onset is likely a consequence of the higher number of smaller droplets that form with more aerosols, suppressing rain formation.
The above insights highlight the utility of our proposed visualization techniques in the analysis of LES simulation data. In the future, we aim to extend this work with further visualization tools that will enable new applications and give us the ability to answer a broader range of questions relating to, for example, entrainment, mass transport, temperature, updraft, and horizontal winds.

\section{Conclusion}
\label{sec:conclusion}



In this study, we have introduced a novel approach to understanding and visualizing droplet size distributions in simulations of warm clouds using Variational Autoencoders (VAEs). By encoding droplet distributions into a compact latent space and representing them through color spectra, we gain valuable insights into the organization and evolution of droplet sizes over time and across different aerosol concentrations. Our findings reveal that while increased aerosol levels delay the onset of precipitation, the evolution of droplet distributions follows a similar pattern. The visualization techniques presented offer powerful tools for efficient and effective analysis of Large Eddy Simulation (LES) data and permit a deeper understanding of cloud microphysics and its impact on weather and climate predictions. Future work will explore additional visualizations to address a broader range of questions related to cloud dynamics and processes.

\clearpage

\section*{Acknowledgements}

The study was supported as part of the Enabling Aerosol–cloud interactions at GLobal convection-permitting scalES (EAGLES) project (project no. 74358) sponsored by the United States Department of Energy (DOE), Office of Science, Office of Biological and Environmental Research (BER), Earth System Model Development (ESMD) program area. The Pacific Northwest National Laboratory (PNNL) is operated for the DOE by the Battelle Memorial Institute under Contract DE-AC05-76RL01830. The research used high-performance computing resources from the PNNL Research Computing, the BER Earth System Modeling program's Compy computing cluster located at PNNL, and resources of the National Energy Research Scientific Computing Center (NERSC), a U.S. Department of Energy Office of Science User Facility located at Lawrence Berkeley National Laboratory, operated under Contract No. DE-AC02-05CH11231, using NERSC awards ALCC-ERCAP0025938 and BER-ERCAP0024471.


\bibliographystyle{ieeetr}
\bibliography{refs}

\end{document}